\documentclass[sigconf,nonacm]{aamas} 

\usepackage{balance} % for balancing columns on the final page
\usepackage{float}
\usepackage{subcaption}
\usepackage{algorithm}
\usepackage{algpseudocode}

\usepackage{booktabs} % 用于创建更美观的表格
\usepackage{multirow} % 用于表格中的多行单元格

\title[AAMAS-2026 Formatting Instructions]{AC-MASAC: An Attentive Curriculum Learning Framework for Heterogeneous UAV Swarm Coordination}

\author{Wanhao Liu}
\affiliation{
  \institution{Guangdong University of Technology} % 请填入单位名称
  \city{Guangzhou}                    % 城市
  \country{China}              % 国家
}
\email{3123006457@mail2.gdut.edu.cn}        % 邮箱

\author{Junhong Dai}
\affiliation{
  \institution{Guangdong University of Technology} % 请填入单位名称
  \city{Guangzhou}                    % 城市
  \country{China}              % 国家
}
\email{2112304096@gdut.edu.cn}

\author{Yixuan Zhang}
\affiliation{
  \institution{Guangdong University of Technology} % 请填入单位名称
  \city{Guangzhou}                  
  \country{China}              % 国家
}
\email{zhangyixuan58@mails.gdut.edu.cn}

\author{Shengyun Yin}
\affiliation{
  \institution{Guangdong University of Technology} % 请填入单位名称
  \city{Guangzhou}                    % 城市
  \country{China}              % 国家
}
\email{2832003829@mails.gdut.edu.cn}

\author{Panshuo Li}
\authornote{Corresponding author.} % 标记通讯作者
\affiliation{
  \institution{Guangdong University of Technology} % 请填入单位名称
  \city{Guangzhou}                    % 城市
  \country{China}              % 国家
}
\email{panshuoli812@gdut.edu.cn}

\begin{abstract}
Cooperative path planning for heterogeneous UAV swarms poses significant challenges for Multi-Agent Reinforcement Learning (MARL), particularly in handling asymmetric inter-agent dependencies and addressing the risks of sparse rewards and catastrophic forgetting during training. To address these issues, this paper proposes an attentive curriculum learning framework (AC-MASAC). The framework introduces a role-aware heterogeneous attention mechanism to explicitly model asymmetric dependencies. Moreover, a structured curriculum strategy is designed, integrating hierarchical knowledge transfer and stage-proportional experience replay to address the issues of sparse rewards and catastrophic forgetting. The proposed framework is validated on a custom multi-agent simulation platform, and the results show that our method has significant advantages over other advanced methods in terms of Success Rate, Formation Keeping Rate, and Success-weighted Mission Time. The code is available at \textcolor{red}{\url{https://github.com/Wanhao-Liu/AC-MASAC}}.
\end{abstract}


\newcommand{\BibTeX}{\rm B\kern-.05em{\sc i\kern-.025em b}\kern-.08em\TeX}

\begin{document}

%%% The following commands remove the headers in your paper. For final 
%%% papers, these will be inserted during the pagination process.

\pagestyle{fancy}
\fancyhead{}

%%% The next command prints the information defined in the preamble.

\maketitle 

%%%%%%%%%%%%%%%%%%%%%%%%%%%%%%%%%%%%%%%%%%%%%%%%%%%%%%%%%%%%%%%%%%%%%%%%

\section{Introduction}

\begin{figure}[h!]
  \centering
  \includegraphics[width=1\columnwidth]{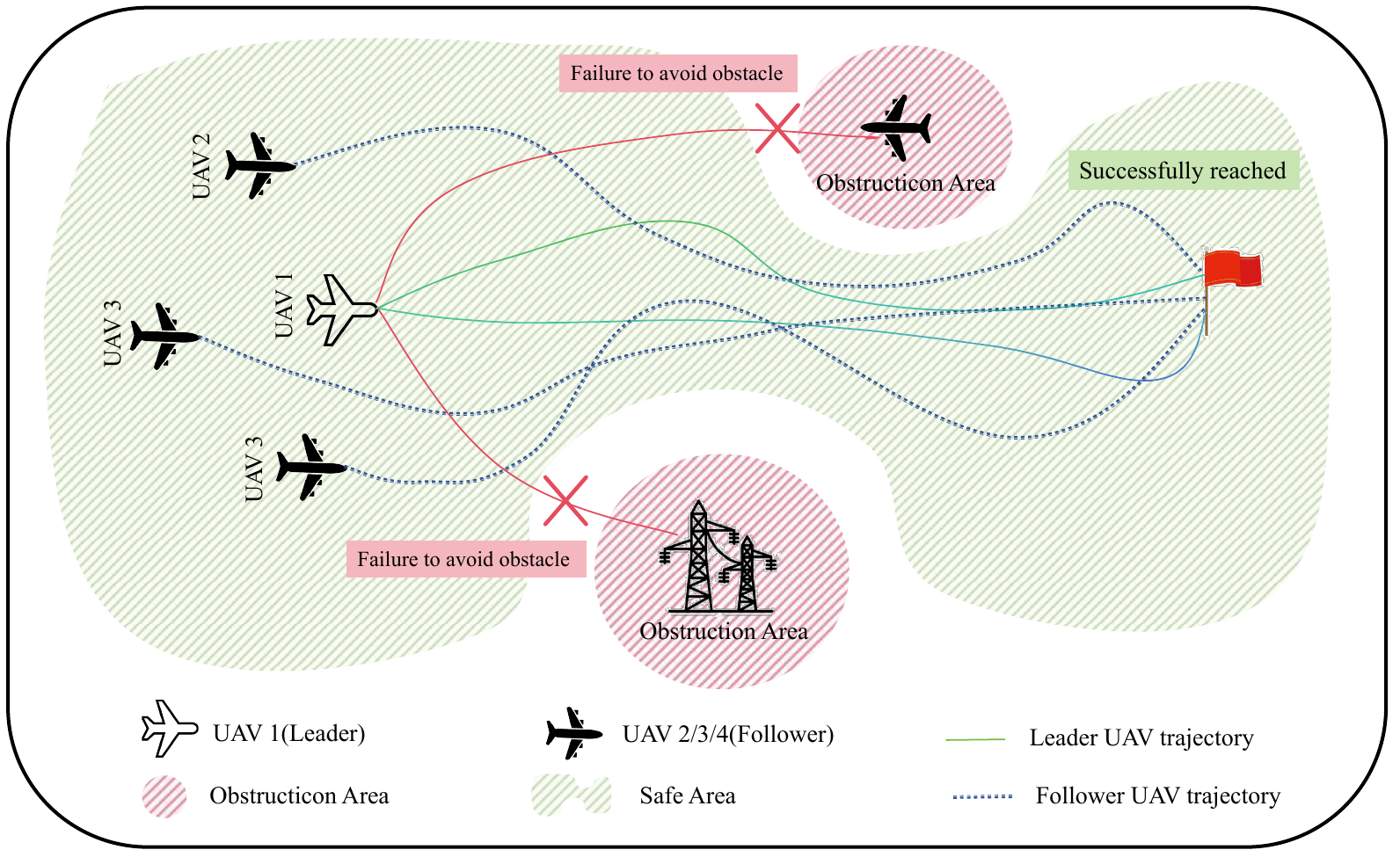}
  \caption{\textbf{Conceptual diagram of the multi-UAV cooperative path planning task.} The swarm, composed of a Leader and multiple Followers, must navigate towards a target while avoiding obstacles and maintaining formation. Each UAV makes decentralized decisions based on its role-specific local observations.}
  \label{fig:task_schematic_diagram}
  \Description{The AC-MASAC autonomous exploration system for heterogeneous UAV swarms. Each agent perceives its local environment through a role-differentiated observation vector—comprising its own kinematic state and critical information about the target, obstacles, or the leader—and makes decentralized decisions through the AC-MASAC algorithm. This enables the swarm to achieve efficient cooperative path planning, robust formation keeping, and effective collision avoidance.}
\end{figure}

Unmanned Aerial Vehicle (UAV) swarms play an increasingly vital role in coordinated tasks such as logistics, inspection, and search-and-rescue \cite{hanover2024autonomous}. Central to these applications is the challenge of cooperative path planning—generating collision-free, dynamically feasible trajectories for all agents to achieve a common goal, as illustrated in Figure~\ref{fig:task_schematic_diagram}. Although this is a widely researched area with numerous existing solutions \cite{wang2022multi,shao2019path,xing2024multi}, current methods still face limitations when addressing specific challenges.

Historically, approaches to this problem can be broadly divided into traditional and learning-based methods. Traditional methods, largely derived from robotics research, include techniques such as sampling-based planners (e.g., RRT* \cite{karaman2011anytime,liu2017dynamic}) and potential field methods (e.g., APF \cite{khatib1986real}). While valuable in certain contexts, these approaches often share common limitations: they typically require a precise model of the environment, exhibit limited adaptability to unforeseen dynamic events, and face significant computational scaling challenges as the swarm size grows.

To overcome the challenges of model dependency and poor adaptability, Multi-Agent Reinforcement Learning (MARL) has emerged as a powerful paradigm for coordinating UAV swarms. Early approaches often treated agents as independent learners (ILs), where each agent learned its policy using standard single-agent RL algorithms like Q-learning or DDPG, considering other agents as part of a dynamic environment \cite{tampuu2017multiagent}. However, this method struggles with the non-stationarity problem, as the concurrent evolution of all agents' policies makes the environment highly unstable and hinders convergence. To address this core challenge, the Centralized Training with Decentralized Execution (CTDE) framework has become the dominant paradigm in the field \cite{foerster2018deep,lowe2017multi}. Within this framework, one prominent line of research focuses on value-decomposition methods. These approaches learn a centralized but factorizable value function, starting with foundational techniques like Value-Decomposition Networks (VDN) \cite{sunehag2017value} and evolving to more expressive methods such as QMIX \cite{rashid2020monotonic} and QTRAN \cite{son2019qtran}. While powerful, these methods are primarily designed for discrete action spaces because selecting an optimal action requires finding the argmax over the joint action space, which becomes an intractable optimization problem in continuous domains. Consequently, Actor-Critic based methods have proven more effective for tasks requiring continuous control. They decouple the problem by learning a dedicated policy network (the actor) that directly maps states to continuous actions, while the critic evaluates the actions provided by the actors. This principle was successfully extended to the multi-agent domain by the Multi-Agent Deep Deterministic Policy Gradient (MADDPG) algorithm \cite{lowe2017multi}, which builds upon DDPG—a method inherently designed for continuous control. Following this line of work, and to address the overestimation bias often found in single-critic methods, the MATD3 algorithm was proposed in \cite{ackermann2019reducing}. It extends the clipped double-Q learning and delayed policy updates from the single-agent TD3 algorithm to the multi-agent setting, leading to more stable value estimates. These foundational Actor-Critic approaches established a powerful and flexible framework for tasks involving continuous control in MARL.

While the above mentioned methods were effective, they often faced challenges with training stability and sample efficiency. To address these issues, the use of Soft Actor-Critic (SAC) \cite{haarnoja2018soft} has been explored in the multi-agent context. The MASAC algorithm, as presented in \cite{fang2024multi,han2025deep}, incorporates the maximum entropy objective to encourage more thorough exploration, leading to more robust and efficient coordination policies. Furthermore, as the research focus shifted towards larger-scale swarms, the challenge of effectively processing and filtering the vast amount of peer information became prominent. To this end, attention mechanisms were integrated into the framework. An attention-based critic was introduced in \cite{iqbal2019actor}, allowing agents to selectively focus on more relevant neighbors when evaluating state-values. Going a step further, attention was applied to the actor network in \cite{sun2024t2mac}, enabling agents to learn a communication strategy to actively select which peers to broadcast their information to. These methods, within a typical homogeneous swarm setting, have significantly improved decision-making in dense and complex scenarios.

Despite these significant advances, applying these frameworks to more complex, structured swarm scenarios reveals two key open challenges. On the one hand, much of the foundational work has centered on homogeneous agent systems, where agents are considered interchangeable. This approach, while broadly applicable, presents limitations when modeling heterogeneous swarms (e.g., Leader-Follower teams), which are characterized by asymmetric inter-agent dependencies that are not explicitly captured by standard models. On the other hand, the risks associated with sparse rewards and catastrophic forgetting remain primary obstacles to achieving stable and efficient policy convergence when applying these algorithms directly to complex tasks. To address these challenges of heterogeneity and training stability, this paper presents the AC-MASAC (Attentive Curriculum-driven Multi-Agent Soft Actor-Critic) algorithm. The contributions of this work are summarized as follows:

\begin{itemize}
    \item We propose a heterogeneous attention model within the MASAC framework that explicitly models the Leader-Follower dynamic. Its role-aware information selection mechanism is designed to capture the asymmetric inter-agent dependencies overlooked by homogeneous approaches.

    \item We design a structured curriculum learning strategy that systematically mitigates the risks of sparse rewards and catastrophic forgetting. By combining stage-proportional experience replay with a hierarchical policy transfer mechanism, our framework enables stable and efficient learning across tasks of increasing complexity.
\end{itemize}

The remainder of this paper is organized as follows. Section 2 presents the problem statement, including the system model and formulation. In Section 3, the proposed AC-MASAC framework is detailed, covering both the heterogeneous actor-critic architecture and the structured curriculum learning strategy. In Section 4, the effectiveness of the proposed method is validated through comprehensive simulation experiments and ablation studies. Finally, Section 5 provides the conclusion.

%%%%%%%%%%%%%%%%%%%%%%%%%%%%%%%%%%%%%%%%%%%%%%%%%%%%%%%%%%%%%%%%%%%%%%%%
\section{PROBLEM STATEMENT}

This paper addresses the problem of cooperative path planning for a heterogeneous multi-UAV swarm operating in a two-dimensional environment with dynamic obstacles. This problem is formally modeled as a Partially Observable Markov Decision Process (POMDP) \cite{kaelbling1998planning}, which provides a rigorous framework for sequential decision-making under uncertainty. The swarm is composed of a designated Leader agent and multiple Follower agents. The primary objective for the Leader is to navigate from a starting position to a target location while avoiding collisions. The primary objective for the Followers is to maintain a specified formation relative to the Leader's trajectory.

To formulate this sequential decision-making problem for a learning-based solution, we define the core components of the agent-environment interface. Each agent $i$ perceives a local observation $z_i$ based on its role. The Leader's observation is a vector $z_i^{(L)} = [x_i, y_i, v_i, \psi_i, x_G, y_G, O_{\text{flag}}]$, which includes its position $(x_i, y_i)$, speed $v_i$, heading angle $\psi_i$, the target's coordinates $(x_G, y_G)$, and a binary obstacle detection flag. In contrast, a Follower's observation is $z_i^{(F)} = [x_i, y_i, v_i, \psi_i, x_L, y_L, v_L]$, which contains its own state alongside the Leader's position $(x_L, y_L)$ and speed $v_L$. The action space for each agent is continuous, where an action $a = [a_c, \omega_c]$ consists of a linear acceleration command $a_c$ and an angular velocity command $\omega_c$, each bounded by the UAV's specific kinematic limits.
The performance of any policy aimed at solving this problem is evaluated using three key metrics. The Success Rate (SR) is defined as the proportion of episodes in which the Leader successfully reaches the target without collision. The Formation Keeping Rate (FKR) measures coordination quality, calculated as the proportion of time steps during which Followers successfully maintain their specified formation constraints. Finally, the Success-weighted Mission Time (SMT), defined as the average episode length multiplied by the Success Rate, is used to evaluate the overall efficiency of the generated solution.

%%%%%%%%%%%%%%%%%%%%%%%%%%%%%%%%%%%%%%%%%%%%%%%%%%%%%%%%%%%%%%%%%%%%%%%%
\begin{figure*}[t]
    \centering
    \includegraphics[width=1.0\linewidth, trim=1cm 1.8cm 2.8cm 1.5cm,clip]{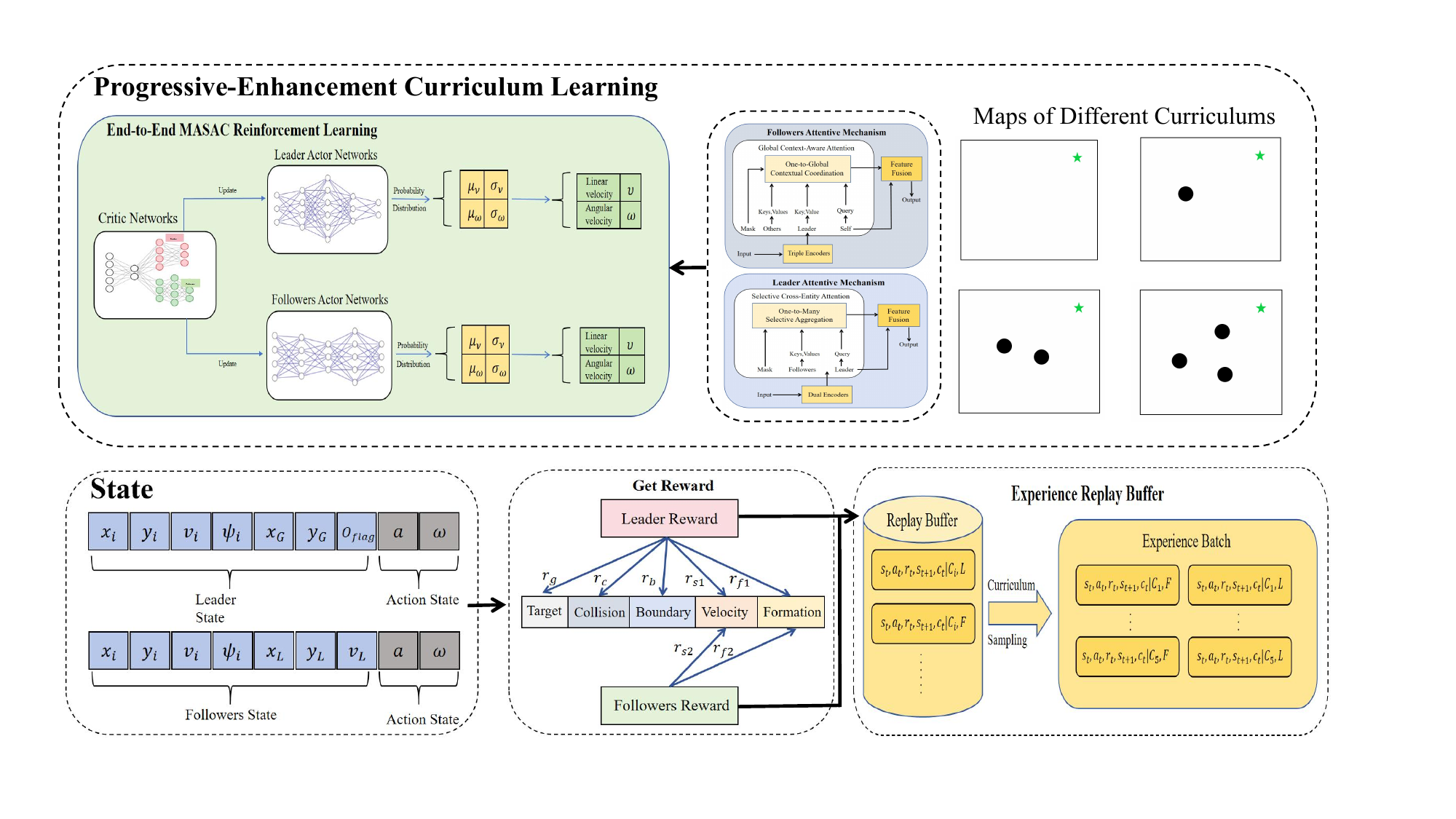}
    \caption{\textbf{AC-MASAC framework overview.} \textit{Top}: Curriculum learning module controlling task difficulty. \textit{Bottom}: Attention-enhanced actor-critic networks processing states for action generation and training via a curriculum-managed replay buffer.}
    \label{fig:all}
    \Description{This figure illustrates the AC-MASAC architecture, a framework where reinforcement learning is guided by a progressive curriculum. The system processes role-differentiated State inputs through Actor-Critic networks enhanced with heterogeneous attention mechanisms to produce actions. These actions generate a multi-component Reward, and the resulting experience is stored in a Replay Buffer. The Curriculum Learning strategy centrally governs both the increasing complexity of the tasks (Maps) and the experience sampling from the buffer, enabling the agent to learn efficiently from simple to complex scenarios.}
\end{figure*}

\section{AC-MASAC Approach}
\subsection{POMDP Formulation}
The cooperative path planning problem under partial observability is modeled as a decentralized Partially Observable Markov Decision Process (POMDP) \cite{ragi2013uav,wang2020multi}, an extension of the traditional Markov Decision Process \cite{bertsekas2019reinforcement,bellman1957markovian}. The POMDP is defined by the tuple $\langle \mathcal{S}, \mathcal{A}, \mathcal{P}, \mathcal{R}, \mathcal{Z}, \mathcal{O}, \gamma \rangle$. Here, $\mathcal{S}$ represents the set of true environmental states, which are not directly observable by the agents. $\mathcal{A}$ is the joint action set composed of individual agent actions, which are known to the agents. $\mathcal{P}$ is the state transition function, governed by the environment dynamics and generally unknown to the agents. $\mathcal{R}$ is the reward function. $\mathcal{Z}$ is the set of joint observations, from which each agent draws its local view. $\mathcal{O}$ is the observation function that maps states to observations, which is also unknown. Finally, $\gamma$ is the discount factor.

\textbf{State Space ($\mathcal{S}$):} Following the Centralized Training with Decentralized Execution (CTDE) paradigm, the global state $s_t \in \mathcal{S}$ is defined as the concatenation of all agents' local observations: $s_t = \{z_t^{(L)}, z_t^{(F_1)}, \dots, z_t^{(F_N)}\}$. This global state serves as the input for the centralized critic to estimate the joint value function based on the swarm's collective perception.

\textbf{Observation Space ($\mathcal{Z}$):} The observation for the $i$-th agent, $z_i \in \mathcal{Z}_i$, consists of its own flight status and role-specific information. For the Leader UAV, the observation vector is $z_i^{(L)} = [x_i,y_i,v_i,\psi_i,x_G,y_G,O_{\text{flag}}]$, where $O_{\text{flag}}$ is a binary indicator set to 1 if the nearest obstacle is within 40m, and 0 otherwise. For a Follower UAV, its observation vector is $z_i^{(F)} = [x_i,y_i,v_i,\psi_i,x_L,y_L,v_L]$, containing information about the Leader. The observation function $\mathcal{O}$ is subject to simulated sensor noise, modeled as an additive Gaussian process $\mathcal{N}(0, \sigma_{\text{noise}}^2)$, where $\sigma_{\text{noise}}=0.1$.

\textbf{Action Space ($\mathcal{A}$):} The action for the $i$-th agent, $a_i \in \mathcal{A}_i$, is a continuous two-dimensional vector $a_i = [u_i, \omega_i]^T$, representing linear acceleration and angular velocity. These actions are constrained by heterogeneous kinematics: the Leader UAV has a velocity range of [10, 20] m/s and acceleration $|u_i| \leq 3 \, \text{m/s}^2$, while Followers have a velocity range of [10, 30] m/s and acceleration $|u_i| \leq 6 \, \text{m/s}^2$.The angular velocity for all agents is constrained to $\omega_i \in [-\pi/3, \pi/3]$ \text{rad/s}.

\textbf{Transition Function ($\mathcal{P}$):} The state transition function $s_{t+1} \sim \mathcal{P}(s_t, a_t)$ is deterministic and governed by the discretized UAV dynamics model\cite{fang2024multi}, with a simulation time step of $\Delta t = 0.1s$:
    \begin{equation}
        \begin{cases}
        x_i(t+1) = x_i(t) + v_i(t) \Delta t \cos \psi_i(t) \\
        y_i(t+1) = y_i(t) + v_i(t) \Delta t \sin \psi_i(t) \\
        v_i(t+1) = v_i(t) + u_i(t) \Delta t \\
        \psi_i(t+1) = \psi_i(t) + \omega_i(t) \Delta t
        \end{cases}
    \label{eq:dynamics}
    \end{equation}
    
\textbf{Reward Function ($\mathcal{R}$):} Role-specific reward functions are designed to guide the agents towards their respective objectives. The total reward for each agent is the sum of several components, and the discount factor is denoted by $\gamma$.

\textbf{Leader's Reward ($R_t^L$):}

    \textbf{Goal Achievement Reward ($r_g$):} A significant sparse reward is provided upon reaching the target. To mitigate the challenge of this sparse signal, a dense reward shaping component based on the negative Euclidean distance to the goal, denoted as $d_g$, is also included.
    \begin{equation}
    r_g = \begin{cases}
      1000.0, & \text{if } d_g < 40 \\
      -0.001 \cdot d_g, & \text{otherwise}
      \end{cases}
    \end{equation}
    
    \textbf{Obstacle Penalty ($r_o$):} To ensure a safe distance from obstacles, a staged penalty is imposed when the Euclidean distance $d_o$ to the nearest obstacle falls below a certain threshold.
    \begin{equation}
      r_o = \begin{cases}
      -500.0, & \text{if } d_o < 20 \\
      -2.0, & \text{if } 20 \le d_o \le 40 \\
      0.0, & \text{otherwise}
      \end{cases}
    \end{equation}
    
    \textbf{Boundary Penalty ($r_e$):} A penalty is applied when the Manhattan distance $d_b$ to the nearest boundary is less than 50.
    \begin{equation}
      r_e = \begin{cases}
      -1.0, & \text{if } d_b < 50 \\
       0.0, & \text{otherwise}
      \end{cases}
    \end{equation}
    
    \textbf{Formation Distance Penalty ($r_{f1}$):} A linear penalty is imposed when the maximum distance to any follower, $\max_{j}(d_{L,j})$, exceeds the ideal formation distance of 50.
        \begin{equation}
          r_{f1} = \begin{cases}
          0.0, & \text{if } \max_{j}(d_{L,j}) \le 50 \\
          -0.001 \cdot \max_{j}(d_{L,j}), & \text{otherwise}
          \end{cases}
        \end{equation}
    
    \textbf{Velocity Matching Reward ($r_{s1}$):} A reward is given for each follower where the speed difference with the leader is less than 1. The total reward is the sum of these individual rewards over all $N$ followers.
    \begin{equation}
    r_{s1} = \sum_{i=1}^{N} \begin{cases}
    1.0, & \text{if } |v_L - v_{F,i}| < 1 \\
    0.0, & \text{otherwise}
    \end{cases}
    \end{equation}
    
The total reward for the Leader is the sum of these components: $R_t^L = r_g + r_o + r_e + r_{f1} + r_{s1}$.

\textbf{Follower's Reward ($R_t^F$):}

    \textbf{Formation Distance Reward ($r_{f2}$):} A staged reward is given when the distance $d_L$ to the leader is within the ideal range; otherwise, a linear penalty is imposed.
    \begin{equation}
      r_{f2} = \begin{cases}
      100.0, & \text{if } 0 < d_L < 40 \\
      10.0, & \text{if } 40 \le d_L < 50 \\
      -0.01 \cdot d_L, & \text{otherwise}
      \end{cases}
    \end{equation}
    
    \textbf{Velocity Matching Reward ($r_{s2}$):} When within the ideal formation distance and the speed difference is less than 1, a velocity matching reward is given.
        \begin{equation}
          r_{s2} = \begin{cases}
          100.0, & \text{if } d_L < 50 \text{ and } |v_L - v_i| < 1 \\
          0.0, & \text{otherwise}
          \end{cases}
        \end{equation}

The total reward for the Follower is the sum of its components: $R_t^F = r_{f2} + r_{s2}$.

An episode terminates when the Leader reaches the target, or when any agent collides with an obstacle.

%%%%%%%%%%%%%%%%%%%%%%%%%%%%%%%%%%%%%%%%%%%%%%%%%%%%%%%%%%%%%%%%%%%%%%%%
\subsection{Heterogeneous Actor-Critic Architecture}
\begin{figure}[h!]
  \centering
  \includegraphics[width=1\columnwidth, trim=0cm 2cm 0cm 2cm,clip]{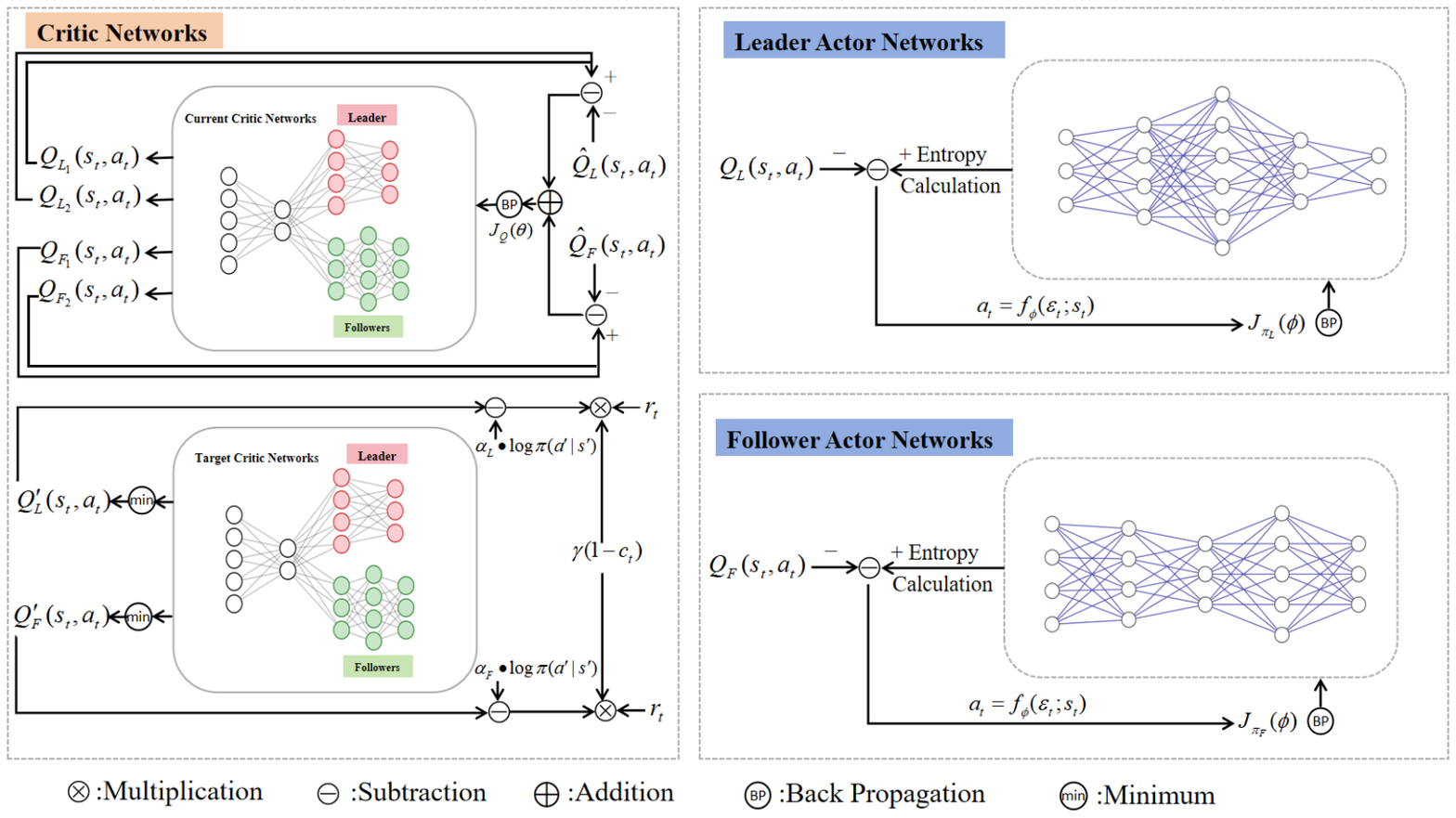}
  \caption{AC-MASAC architecture diagram.}
  \label{fig:ac-masac_arch}
  \Description{This figure details the AC-MASAC's neural network architecture, which adapts the Soft Actor-Critic (SAC) framework for a heterogeneous multi-agent system. The Critic Networks (left) employ a twin Q-network design with corresponding target networks to mitigate overestimation bias, learning separate Q-values for the Leader and Followers by minimizing the soft Bellman error. The Actor Networks (right) are role-differentiated, with distinct policies for the Leader and Followers. Each actor is updated via backpropagation to maximize both the Q-value provided by the critic and its own policy entropy, encouraging efficient exploration and robust performance.}
\end{figure}

The AC-MASAC algorithm is built upon the Soft Actor-Critic (SAC) framework \cite{haarnoja2018soft}, which optimizes a policy $\pi$ to maximize an entropy-regularized objective function. The objective is to find a policy that balances the maximization of cumulative reward with the maximization of its own entropy, which encourages exploration and enhances robustness:
\begin{equation}
\label{eq:sac_objective}
J(\pi) = \sum_{t=0}^{T} \mathbb{E}_{(s_t, a_t) \sim \rho_{\pi}} \left[ r(s_t, a_t) + \alpha \mathcal{H}(\pi(\cdot|s_t)) \right]
\end{equation}
The temperature parameter, $\alpha$, serves as a trade-off coefficient. Instead of being a fixed hyperparameter, $\alpha$ is learned automatically by minimizing the following loss function, which aims to maintain a target entropy level $\bar{\mathcal{H}}$:
\begin{equation}
\label{eq:alpha_loss}
J(\alpha) = \mathbb{E}_{a_t \sim \pi_t} [-\alpha \log \pi_t(a_t|s_t) - \alpha \bar{\mathcal{H}}]
\end{equation}

The core of our contribution lies in the design of a heterogeneous actor-critic architecture, termed AC-MASAC. Its overall framework is depicted in Figure~\ref{fig:all}, while the detailed network architecture is shown in Figure~\ref{fig:ac-masac_arch}. This approach explicitly models the Leader-Follower roles, directly addressing the limitations of the homogeneity assumption identified in the introduction. Instead of using a single, monolithic policy structure, we introduce two distinct, role-specific actor networks and a critic whose structure mirrors this heterogeneity. This design provides an inductive bias that facilitates the learning of specialized, coordinated behaviors, overcoming the challenge of unstructured state representations in prior multi-agent actor-critic methods.

\begin{figure}[h!]
    \centering
    % Subfigure for Leader Actor
    \begin{subfigure}{\columnwidth}
        \centering
        \includegraphics[width=1.1\columnwidth, trim=0cm 13cm 13cm 0cm,clip]{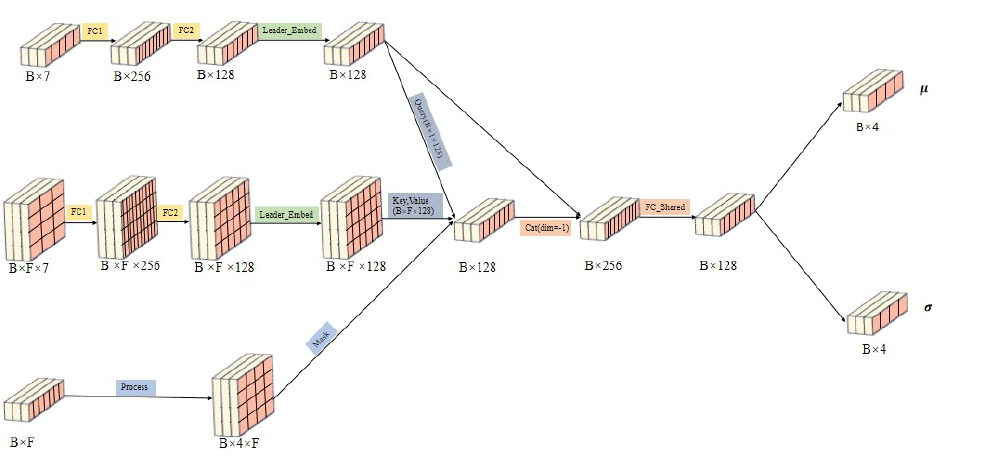}
        \caption{Leader Actor network architecture}
        \label{fig:leader-attn}
    \end{subfigure}
    
    \vspace{1em} % Adds vertical space between subfigures for clarity
    
    % Subfigure for Follower Actor
    \begin{subfigure}{\columnwidth}
        \centering
        \includegraphics[width=1.1\columnwidth, trim=0cm 13cm 13cm 0cm,clip]{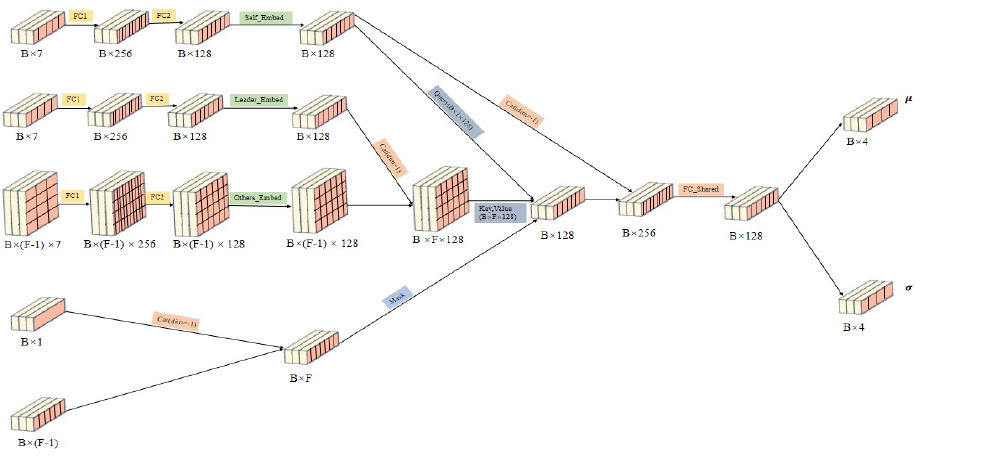}
        \caption{Follower Actor network architecture}
        \label{fig:follower-attn}
    \end{subfigure}
    
    \vspace{1em} % Adds vertical space between subfigures for clarity

    % Subfigure for Critic
    \begin{subfigure}{\columnwidth}
        \centering
        \includegraphics[width=1\columnwidth, trim=0cm 15cm 17cm 0cm,clip]{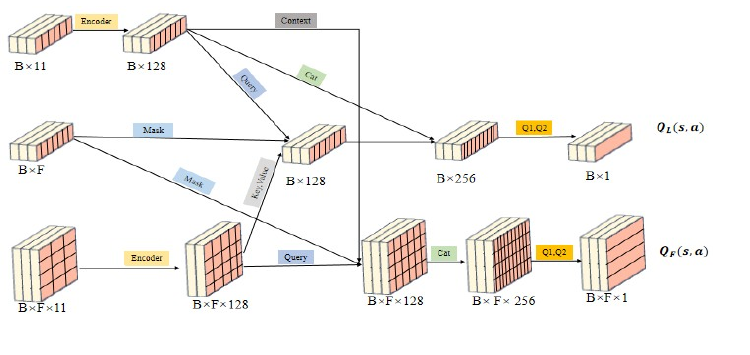}
        \caption{Structured Attention Critic network architecture}
        \label{fig:critic-attn}
    \end{subfigure}
    \caption{\textbf{Attention-based network architectures.} (a) Leader Actor, (b) Follower Actor, and (c) Structured Attention Critic. Inputs are role-specific states; outputs are Gaussian policy parameters ($\mu, \sigma$).}
    \label{fig:attention-architectures}
    \Description{A set of three vertically arranged diagrams detailing the neural network architectures for the AC-MASAC algorithm. Subfigure (a) illustrates the Leader Actor network, which uses selective cross-entity attention to process its own state and the states of all followers, outputting policy parameters mu and sigma. Subfigure (b) shows the Follower Actor network, featuring a triple-encoder design to separately process self, leader, and other follower states before applying global context-aware attention. Subfigure (c) depicts the Structured Attention Critic network, which mirrors the actor patterns in two separate branches to calculate Q-values for both the leader and followers based on their respective state-action pair inputs.}
\end{figure}

Heterogeneous Actor Networks as illustrated in Figure~\ref{fig:attention-architectures}, the architecture employs two distinct actor networks, whose designs are tailored to the specific informational needs of each role.

The Leader Actor, whose architecture is detailed in Figure~\ref{fig:attention-architectures}(a), utilizes a selective cross-entity attention mechanism. The input is the Leader's local observation $z_i^{(L)}$. It is first passed through an embedding layer $\phi_e^L$ to produce an embedding $e^L$, which serves as the query ($Q$). The observations of the followers $\{z_j^{(F)}\}$ are similarly embedded via a shared embedding layer $\phi_e^F$ to produce keys ($K$) and values ($V$). To handle a variable number of followers and partial observability, a head-wise attention mask is applied, which allows the network to ignore non-existent or out-of-range agents during the attention score calculation. The attention mechanism then computes a context vector representing the aggregated follower states.

The Follower Actor(Figure~\ref{fig:attention-architectures}(b)) uses a global context-aware attention mechanism. Its local observation $z_i^{(F)}$ is embedded via $\phi_e^F$ to form the query ($Q$). The context is formed by the Leader's observation $z_i^{(L)}$ and all other followers' observations $\{z_j^{(F)}\}_{j \neq i}$, which are embedded to provide the keys ($K$) and values ($V$). To improve sample efficiency, the Follower Actor networks share parameters across all follower agents.

For both actor networks, the final processed features are projected to produce the parameters of a Gaussian policy, yielding a mean $\mu$ and a log standard deviation $\log\sigma$ that define the distribution from which continuous actions are sampled.

Structured Attention Critic to maintain policy-value consistency within the CTDE framework, the critic's architecture, shown in Figure~\ref{fig:attention-architectures}(c), mirrors the actor's heterogeneous attention structure. It employs a twin Q-network design to mitigate overestimation bias. Each Q-network contains two parallel branches: a \textit{leader branch} that computes the Q-value for the Leader by applying selective attention over follower state-action pairs, and a \textit{follower branch} that computes Q-values for followers using global context attention. During training, the critic receives the global state $s_t$ (approximated by the set of all observations $\{z_t^i\}$) and the joint action $a_t$ to learn a centralized value function. The target Q-value, used for the Bellman update, is computed as the minimum of the two target Q-network outputs. The target networks are updated via a soft update from their main network counterparts: $\theta^{-} \leftarrow \tau\theta+(1-\tau)\theta^{-}$ with a small update rate $\tau$.

For the implementation, learning rates were set to $1 \times 10^{-4}$ for the actor networks and $3 \times 10^{-4}$ for the critic networks. The discount factor $\gamma$ was 0.99, the experience buffer size was $5 \times 10^4$, and the batch size was 256. The soft update rate was $\tau=0.01$. The Adam optimizer was used for all networks. Within each Multi-Head Attention block, the embedding dimension was $E=128$, the number of heads was $H=4$, and a dropout probability of 0.1 was applied to prevent overfitting.

\subsection{Structured Curriculum Learning Framework}
\begin{figure}[H]
  \centering
  \includegraphics[width=1\columnwidth, trim=0cm 14cm 1cm 2cm,clip]{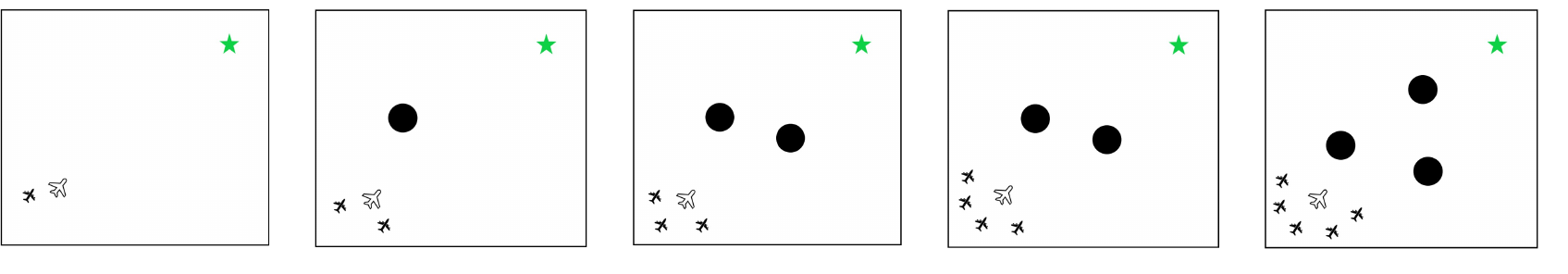}
  \caption{Environment for different levels of training}
  \label{fig:curriculum-stages}
  \Description{A set of different training and testing environments for the curriculum learning stages, in which a UAV swarm (a leader and multiple followers) is at the bottom-left and a green star-shaped target is at the top-right. The complexity of the task progressively increases from left to right: the first stage shows one leader, one follower, and zero obstacles. Subsequent stages show an increasing number of followers and black circular obstacles, culminating in a complex scenario with multiple agents and obstacles.}
\end{figure}
The AC-MASAC algorithm incorporates a curriculum learning framework to structure the training process. The agent is trained on a predefined sequence of tasks, $\mathcal{C} = (T_1, T_2, \dots, T_K)$, ordered by progressively increasing complexity. Each task $T_k$ is defined by a set of environmental parameters $\xi_k = (N_{L,k}, N_{F,k}, N_{O,k})$, representing the number of leaders, followers, and obstacles, respectively, where complexity increases such that $N_{F,k+1} \ge N_{F,k}$ and $N_{O,k+1} \ge N_{O,k}$. To foster policy generalization and prevent overfitting to specific initial configurations, the starting conditions for each episode are randomized. This includes the initial states of the leader and all follower agents, as well as the locations of the goal and all dynamic obstacles, which are sampled from predefined distributions within the operational area. This systematic escalation in environmental difficulty ensures the agent builds a foundation of skills before confronting the full complexity of the final task, as illustrated in Figure~\ref{fig:curriculum-stages}.

\begin{algorithm}
\caption{Progressive Curriculum Learning for AC-MASAC}
\label{alg:ac-masac}
\begin{algorithmic}[1]
\State Initialize MASAC controller $M$, curriculum manager $CM$, stage $s \gets 1$, knowledge transfer $KT \gets \varnothing$
\For{$s = 1$ to $\mathrm{max\_stages}$}
    \State $\mathrm{Task}_s \gets \mathrm{generate\_fixed\_task}(\mathrm{difficulty}_s)$
    \If{$s > 1$}
        \State $M \gets KT.\mathrm{transfer}(M, \mathrm{agent\_count}_s)$
    \EndIf
    \State $\mathrm{success\_rate} \gets 0$, $\mathrm{episode} \gets 0$
    \While{$\mathrm{success\_rate} < \theta$ and $\mathrm{episode} < \mathrm{max\_episodes}$}
        \State $E \gets \mathrm{Task}_s.\mathrm{create\_environment}()$
        \State $\mathrm{ratio} \gets \mathrm{adaptive\_sampling\_ratio}(s, \mathrm{episode})$
        \State $M.\mathrm{set\_replay\_ratio}(\mathrm{ratio})$
        \State $M.\mathrm{train}(\mathrm{current\_stage}=s, \mathrm{sampling\_ratio}=\mathrm{ratio})$
        \State $\mathrm{success\_rate} \gets \mathrm{evaluate\_performance}(M, E)$
        \State $\mathrm{episode} \gets \mathrm{episode} + 1$
    \EndWhile
    \State $KT \gets M.\mathrm{export\_parameters}()$
\EndFor
\end{algorithmic}
\end{algorithm}
The sequencing of tasks and the criteria for switching between them follow a deterministic linear progression through the task sequence $\mathcal{C}$. The transition from task $T_k$ to $T_{k+1}$ is triggered only when the agent's performance, evaluated over a sliding window of $W$ episodes, meets a set of predefined thresholds. These criteria are based on the metrics defined in Section 2: a success rate $SR \ge \theta_{SR}$, a reward coefficient of variation $CV_R \le \theta_{CV}$, and a formation keeping rate $FKR \ge \theta_{FKR}$.

A hierarchical knowledge transfer mechanism is applied to the network parameters to leverage previously learned policies upon switching to a new task $T_{k+1}$. Let $\theta_A$ and $\phi_C$ denote the parameters of the actor and critic networks, respectively. Specifically, the Leader's actor policy is fully transferred ($\theta_{A,L}^{(k+1)} \leftarrow \theta_{A,L}^{(k)}$), as its core objective remains consistent across tasks. For the Follower actors, parameters for the $N_{F,k}$ existing agents are preserved ($\theta_{A, F_i}^{(k+1)} \leftarrow \theta_{A, F_i}^{(k)}$ for $i=1, \dots, N_{F,k}$), while any newly introduced followers ($j > N_{F,k}$) have their parameters replicated from the most experienced existing follower ($\theta_{A, F_j}^{(k+1)} \leftarrow \theta_{A, F_{N_{F,k}}}^{(k)}$) to bootstrap coordination behavior. In contrast, the critic networks are re-initialized ($\phi_C^{(k+1)} \sim \text{Init()}$), allowing the value function to adapt to the new task's specific dynamics and reward landscape without being biased by outdated value estimations.

To mitigate catastrophic forgetting, a stage-proportional experience sampling mechanism manages the experience replay buffer $\mathcal{D}$ as a collection of stage-specific sub-buffers, $\mathcal{D} = \bigcup_{j=1}^{k} \mathcal{D}_j$. During training at stage $k$, the proportion of experiences sampled from historical stages ($j<k$), denoted $\rho_{\text{old}}(k)$, is determined by a stage-dependent decay function:
\begin{equation}
\rho_{\text{old}}(k) = \max(\rho_{\text{base}} - \eta \cdot (k-1), \rho_{\text{min}})
\label{eq:ratio_symbolic}
\end{equation}
where $\rho_{\text{base}}$ is the initial proportion of historical experiences, $\eta$ is the per-stage decay rate, and $\rho_{\text{min}}$ is the minimum proportion. A training mini-batch is thus composed of samples drawn from the current stage buffer $\mathcal{D}_k$ and the historical buffer $\mathcal{D}_{\text{hist}} = \bigcup_{j=1}^{k-1} \mathcal{D}_j$ according to this ratio. This mechanism maintains a balance between knowledge retention and adaptation to the current task. In our implementation, the hyperparameters are set as follows: $\theta_{SR}=0.9$, $\theta_{CV}=0.3$, $\theta_{FKR}=0.8$, $W=15$, $\rho_{\text{base}}=0.8$, $\eta=0.1$, and $\rho_{\text{min}}=0.2$.

\section{EXPERIMENTS AND DISCUSSION}
To comprehensively evaluate the performance of the proposed AC-MASAC algorithm, we compared it with two mainstream baseline methods. For the non-learning-based approach, we utilized RRT* \cite{karaman2011anytime} combined with a Pure Pursuit Controller, which uses complete global information to generate a centralized reference path, serving as a static planning benchmark. For learning-based methods, we chose MASAC \cite{fang2024multi} and MADDPG \cite{lowe2017multi} as comparative baselines. All hyperparameters for the baseline algorithms were set to the default recommended values from their official open-source implementations.
\begin{figure}[h!]
  \centering
  \begin{subfigure}{0.15\textwidth}
    \centering
    \includegraphics[width=\linewidth, trim=0cm 14cm 24cm 2cm,clip]{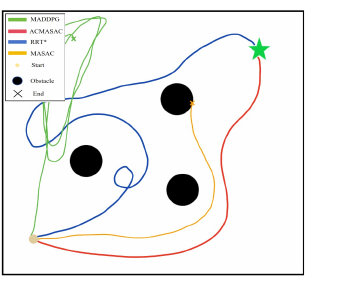}
    \caption{Experiment 1}
    \label{fig:experiment1}
  \end{subfigure}
  \hfill
  \begin{subfigure}{0.15\textwidth}
    \centering
    \includegraphics[width=\linewidth, trim=0cm 14cm 24cm 2cm,clip]{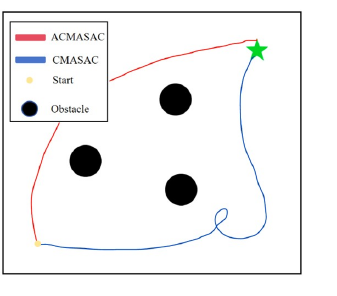}
    \caption{Experiment 2}
    \label{fig:experiment2}
  \end{subfigure}
  \hfill
  \begin{subfigure}{0.15\textwidth}
    \centering
    \includegraphics[width=\linewidth, trim=0cm 14cm 24cm 2cm,clip]{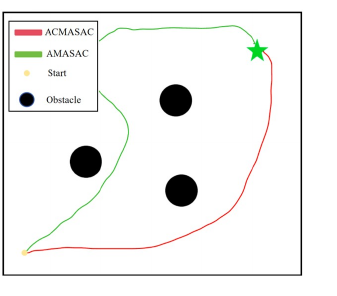}
    \caption{Experiment 3}
    \label{fig:experiment3}
  \end{subfigure}
  \hfill
  \centering
  \begin{subfigure}{0.15\textwidth}
    \centering
    \includegraphics[width=\linewidth, trim=0cm 3.5cm 27cm 0cm,clip]{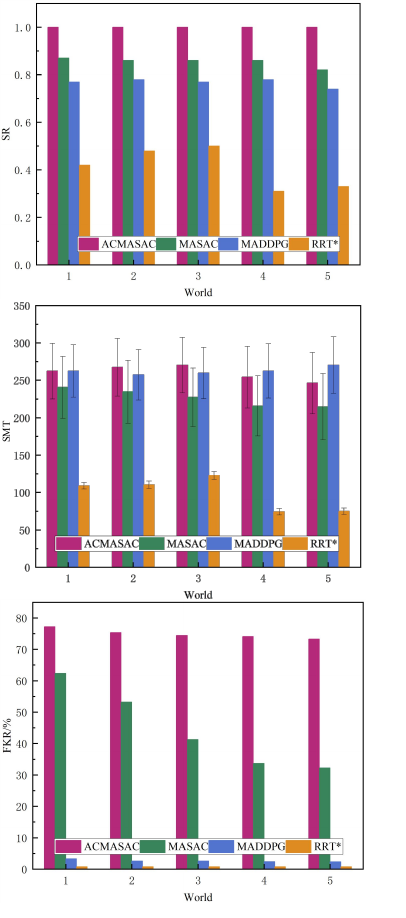}
    \caption{Main Comparison}
    \label{fig:sim1}
  \end{subfigure}
  \hfill
  \begin{subfigure}{0.15\textwidth}
    \centering
    \includegraphics[width=\linewidth, trim=0cm 3.5cm 27cm 0cm,clip]{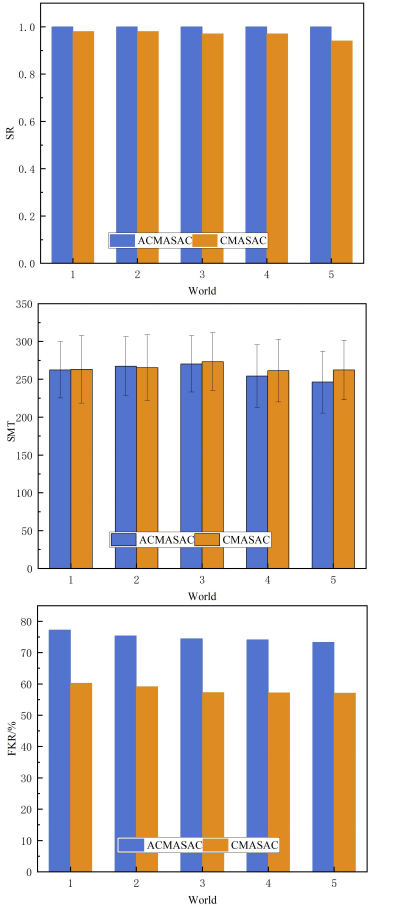}
    \caption{Ablation 1}
    \label{fig:sim2}
  \end{subfigure}
  \hfill
  \begin{subfigure}{0.15\textwidth}
    \centering
    \includegraphics[width=\linewidth, trim=0cm 3.5cm 27cm 0cm,clip]{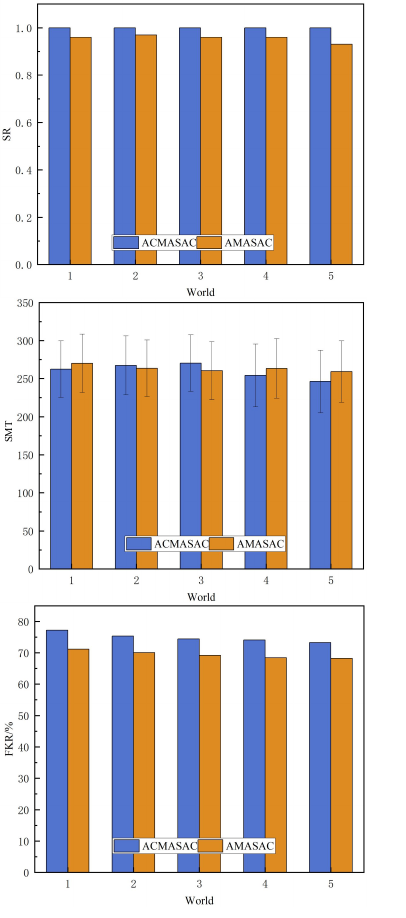}
    \caption{Ablation 2}
    \label{fig:sim3}
  \end{subfigure}
  \caption{Experimental trajectory diagram, SR, SMT and FKR, (a)-(c) show the leader trajectory plots from three different experiments in simulation environment world5. (d)-(f) show Performance Results plots in various test worlds, including SR, SMT and FKR. Histogram means the average value for different metrics and the error bar is the range under a 95\% confidence interval. }
  \label{fig:all_simulations}
  \Description{Three simulation results showing performance comparison across different algorithms}
\end{figure}
\subsection{Simulation Experiment}

The simulation environment, custom-developed based on the OpenAI Gym framework with Pygame for visualization, simulates a $700 \times 600$-meter airspace and is designed to model heterogeneous agent dynamics and dynamic obstacle configurations.

To evaluate generalization, a suite of test environments with varying difficulty levels was used, as depicted in Figure~\ref{fig:curriculum-stages}. Each algorithm was tested for 100 episodes per environment, with each episode lasting a maximum of 1000 iterations and initiated with randomized agent positions, target locations, and obstacle placements. The performance of four algorithms AC-MASAC, MASAC, MADDPG, and RRT*—were recorded based on three key metrics: Success Rate (SR), Success-weighted Mission Time (SMT), and Formation Keeping Rate (FKR).

As shown in the representative trajectories in Figure~\ref{fig:all_simulations}(a), AC-MASAC consistently generated a smooth and direct path to the target, benefiting from its attention-based coordination and curriculum-honed policy. The MADDPG algorithm often produced hesitant or circuitous routes, indicating struggles with multi-agent credit assignment. The MASAC algorithm, while better than MADDPG, still showed inefficiencies in obstacle-dense areas. The RRT* algorithm, being a non-learning method, found feasible paths but they were frequently suboptimal and lacked the fluid dynamics of the learned policies. 

The quantitative results in Figure~\ref{fig:all_simulations}(d) further support these observations. Across all test worlds, AC-MASAC achieved the highest SR and, most critically, a vastly superior FKR, demonstrating its robust coordination capabilities. While MASAC and MADDPG showed reasonable performance in simpler worlds, their success rates and formation stability degraded significantly as complexity increased, indicating poorer generalization. The RRT* method, while consistent, scored lowest on efficiency (SMT) and had a near-zero FKR as it lacks an inherent coordination mechanism. Overall, while all learning-based methods demonstrated flexibility, AC-MASAC exhibited the best all-around performance. The significant lead in FKR explicitly validates the effectiveness of our heterogeneous attention mechanism in modeling Leader-Follower dynamics, while the superior SR and SMT in complex environments underscore the contribution of the structured curriculum framework in achieving robust policy convergence where other methods falter.

\subsection{Ablation Experiment}
\subsubsection*{1) Attention Mechanism}
To validate the efficacy of the proposed attention mechanism, we conducted an ablation study comparing our final AC-MASAC algorithm against a variant without the attention mechanism, hereafter referred to as C-MASAC (Curriculum-only MASAC). Both models were trained and tested under identical settings.

As illustrated in the representative path comparison in Figure~\ref{fig:all_simulations}(b), AC-MASAC navigates around obstacles while maintaining a smooth and goal-oriented trajectory. In contrast, C-MASAC, lacking dynamic awareness of teammate states, follows a path that appears more tortuous and hesitant, particularly at critical junctures requiring collective decision-making. AC-MASAC effectively leverages the attention mechanism to dynamically weigh the importance of information from different teammates, leading to more cooperative decisions that directly manifest as tighter formations and more optimal path choices.

Furthermore, the quantitative metrics in Figure~\ref{fig:all_simulations}(e) show that AC-MASAC's Formation Keeping Rate (FKR) consistently surpassed that of C-MASAC by an average of over 15\% across all test environments. This provides strong evidence for the central role of the attention mechanism in enhancing multi-agent coordination. Concurrently, the higher Success Rate (SR) indicates that improved coordination translates directly into a greater probability of mission success.

\subsubsection*{2) Curriculum Learning}
To evaluate the role of curriculum learning in enhancing algorithm generalization and final performance convergence, we conducted a comparative experiment between our full AC-MASAC algorithm and A-MASAC (Attention-only MASAC), an attention-based variant trained directly in the most complex environment without the structured curriculum.
\begin{figure}[h!]
    \centering
    \includegraphics[width=1.0\linewidth, trim=0cm 0cm 2cm 0cm,clip]{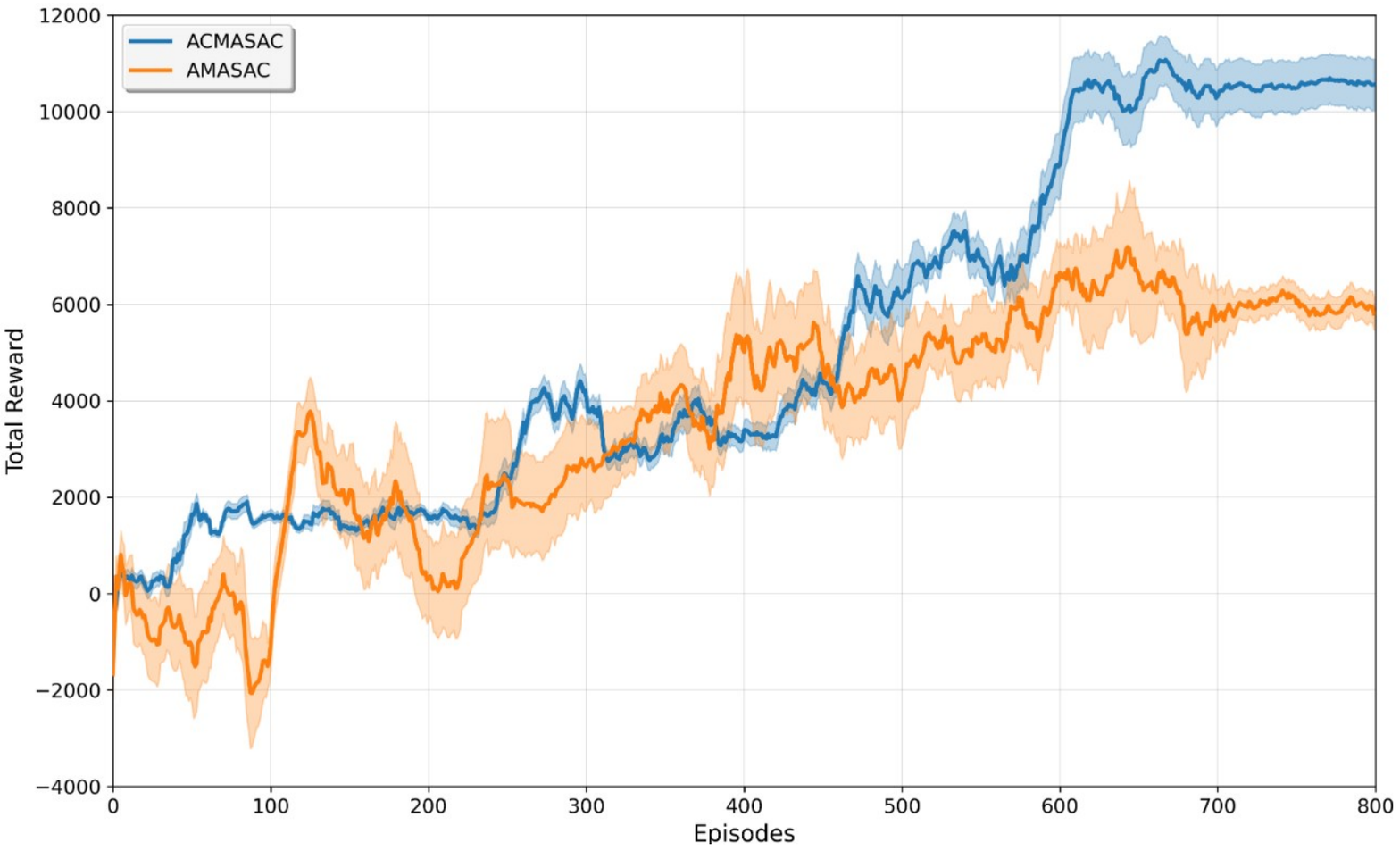}
    \caption{Comparison of training reward curves for AC-MASAC and A-MASAC.}
    \label{fig:reward-curve-cl-ablation}
    \Description{A line graph plotting Total Reward versus Episodes, comparing the performance of AC-MASAC (blue) and A-MASAC (orange). The AC-MASAC curve shows a stable and consistent rise, converging to a high reward of over 10,000. In contrast, the A-MASAC curve exhibits greater volatility and converges to a much lower reward level, around 6,000.}
\end{figure}
As shown in the reward curve in Figure~\ref{fig:reward-curve-cl-ablation}, A-MASAC experienced significant reward fluctuations during the initial training phase due to the immediate exposure to a complex task, and its final converged reward level was notably lower than that of AC-MASAC. This suggests that learning a complex multi-agent coordination policy from scratch is exceedingly difficult.

Conversely, AC-MASAC, through its simple-to-complex curriculum learning paradigm, exhibited a more stable learning curve and converged to a significantly higher performance level, surpassing A-MASAC's final average reward by approximately 82\%. This highlights the importance of curriculum learning in establishing foundational policies and avoiding local optima. The path comparisons and quantitative metrics in Figures~\ref{fig:all_simulations}(c) and~\ref{fig:all_simulations}(f) further corroborate these findings: AC-MASAC consistently outperformed A-MASAC in both Success Rate (SR) and Formation Keeping Rate (FKR), indicating that the knowledge acquired through curriculum learning generalizes more effectively, enabling superior performance in a variety of complex environments.

\subsubsection*{3) Robustness to Communication Imperfections}
While the primary experiments presented above assume ideal communication channels, real-world deployment of UAV swarms inevitably involves non-ideal conditions such as packet loss and latency. To evaluate the robustness of the proposed AC-MASAC against such practical challenges, we conducted additional tests introducing stochastic packet dropout during the decentralized execution phase. In these scenarios, each agent fails to receive state updates from a specific teammate with a predefined probability $p_{\text{drop}}$ at each time step. Qualitative results derived from complex test environments indicate that AC-MASAC maintains resilient coordination performance under moderate packet loss rates (e.g., $p_{\text{drop}} \approx 10-20\%$). Although formation precision naturally exhibits graceful degradation as $p_{\text{drop}}$ increases beyond this range, the swarm successfully avoids catastrophic failures such as inter-agent collisions or complete formation breakup. This observed robustness can be largely attributed to the inherent properties of the proposed heterogeneous attention mechanism, which dynamically learns to down-weight unreliable or missing signals and refocus on available critical information, thereby sustaining effective cooperative decision-making under uncertainty.
%%%%%%%%%%%%%%%%%%%%%%%%%%%%%%%%%%%%%%%%%%%%%%%%%%%%%%%%%%%%%%%%%%%%%%%%

\section{Conclusion}
This paper addresses the cooperative path planning control of a heterogeneous UAV swarm using a multi-agent reinforcement learning method. By defining a Leader-Follower role structure, we established a decentralized Partially Observable Markov Decision Process (POMDP) that takes into account both individual agent dynamics and multi-agent coordination objectives. A heterogeneous actor-critic architecture was developed to address the role-specific decision-making requirements. Within this architecture, a role-aware attention mechanism was designed to process the unstructured state information inherent in multi-agent systems. The constructed system learns distinct policies for Leader and Follower agents based on their differentiated observation and action spaces.

To address the challenges of training instability and slow convergence in complex, sequential tasks, a structured curriculum learning framework was integrated. Based on a sequence of tasks with progressively increasing complexity, the framework utilizes a hierarchical knowledge transfer mechanism for policy initialization and a stage-proportional experience replay strategy to mitigate catastrophic forgetting. Based on this framework, a robust, curriculum-driven policy was synthesized, which adapts to environments of varying difficulty.

The proposed AC-MASAC method is applicable to cases with heterogeneous agent roles and complex, dynamic environments, and it achieves superior performance in success rate and formation keeping compared with conventional MARL baselines. In the future, the AC-MASAC framework will be extended to address the challenges of sim-to-real transfer for deployment on physical hardware platforms.

%%%%%%%%%%%%%%%%%%%%%%%%%%%%%%%%%%%%%%%%%%%%%%%%%%%%%%%%%%%%%%%%%%%%%%%%

%%% The acknowledgments section is defined using the "acks" environment
%%% (rather than an unnumbered section). The use of this environment 
%%% ensures the proper identification of the section in the article 
%%% metadata as well as the consistent spelling of the heading.

%\begin{acks}
%If you wish to include any acknowledgments in your paper (e.g., to 
%people or funding agencies), please do so using the `\texttt{acks}' 
%environment. Note that the text of your acknowledgments will be omitted
%if you compile your document with the `\texttt{anonymous}' option.
%\end{acks}

%%%%%%%%%%%%%%%%%%%%%%%%%%%%%%%%%%%%%%%%%%%%%%%%%%%%%%%%%%%%%%%%%%%%%%%%

%%% The next two lines define, first, the bibliography style to be 
%%% applied, and, second, the bibliography file to be used.

\bibliographystyle{ACM-Reference-Format} 
\bibliography{reference}

@article{wang2022multi,
  title={Multi-robot path planning with due times},
  author={Wang, Hanfu and Chen, Weidong},
  journal={IEEE Robotics and Automation Letters},
  volume={7},
  number={2},
  pages={4829--4836},
  year={2022},
  publisher={IEEE}
}

@article{shao2019path,
  title={Path planning for multi-UAV formation rendezvous based on distributed cooperative particle swarm optimization},
  author={Shao, Zhuang and Yan, Fei and Zhou, Zhou and Zhu, Xiaoping},
  journal={Applied Sciences},
  volume={9},
  number={13},
  pages={2621},
  year={2019},
  publisher={MDPI}
}

@article{xing2024multi,
  title={Multi-UAV adaptive cooperative formation trajectory planning based on an improved MATD3 algorithm of deep reinforcement learning},
  author={Xing, Xiaojun and Zhou, Zhiwei and Li, Yan and Xiao, Bing and Xun, Yilin},
  journal={IEEE Transactions on Vehicular Technology},
  volume={73},
  number={9},
  pages={12484--12499},
  year={2024},
  publisher={IEEE}
  }

@article{liu2017dynamic,
  title={Dynamic path planning based on an improved RRT algorithm for RoboCup robot},
  author={Liu, Chengju and Han, Junqiang and An, Kang},
  journal={Robot},
  volume={39},
  number={1},
  pages={8--15},
  year={2017}
}

@article{khatib1986real,
  title={Real-time obstacle avoidance for manipulators and mobile robots},
  author={Khatib, Oussama},
  journal={The international journal of robotics research},
  volume={5},
  number={1},
  pages={90--98},
  year={1986},
  publisher={Sage Publications Sage CA: Thousand Oaks, CA}
}

@article{hanover2024autonomous,
  title={Autonomous drone racing: A survey},
  author={Hanover, Drew and Loquercio, Antonio and Bauersfeld, Leonard and Romero, Angel and Penicka, Robert and Song, Yunlong and Cioffi, Giovanni and Kaufmann, Elia and Scaramuzza, Davide},
  journal={IEEE Transactions on Robotics},
  volume={40},
  pages={3044--3067},
  year={2024},
  publisher={IEEE}
}

@phdthesis{foerster2018deep,
  title={Deep multi-agent reinforcement learning},
  author={Foerster, J},
  year={2018},
  school={University of Oxford}
}

@inproceedings{iqbal2019actor,
  title={Actor-attention-critic for multi-agent reinforcement learning},
  author={Iqbal, Shariq and Sha, Fei},
  booktitle={International conference on machine learning},
  pages={2961--2970},
  year={2019},
  organization={PMLR}
}

@inproceedings{haarnoja2018soft,
  title={Soft actor-critic: Off-policy maximum entropy deep reinforcement learning with a stochastic actor},
  author={Haarnoja, Tuomas and Zhou, Aurick and Abbeel, Pieter and Levine, Sergey},
  booktitle={International conference on machine learning},
  pages={1861--1870},
  year={2018},
  organization={Pmlr}
}

@inproceedings{karaman2011anytime,
  title={Anytime motion planning using the RRT},
  author={Karaman, Sertac and Walter, Matthew R and Perez, Alejandro and Frazzoli, Emilio and Teller, Seth},
  booktitle={2011 IEEE international conference on robotics and automation},
  pages={1478--1483},
  year={2011},
  organization={ieee}
}

@article{lowe2017multi,
  title={Multi-agent actor-critic for mixed cooperative-competitive environments},
  author={Lowe, Ryan and Wu, Yi I and Tamar, Aviv and Harb, Jean and Abbeel, Pieter and Mordatch, Igor},
  journal={Advances in neural information processing systems},
  volume={30},
  year={2017}
}

@article{ragi2013uav,
  title={UAV path planning in a dynamic environment via partially observable Markov decision process},
  author={Ragi, Shankarachary and Chong, Edwin KP},
  journal={IEEE Transactions on Aerospace and Electronic Systems},
  volume={49},
  number={4},
  pages={2397--2412},
  year={2013},
  publisher={IEEE}
}

@article{wang2020multi,
  title={Multi-agent deep reinforcement learning-based trajectory planning for multi-UAV assisted mobile edge computing},
  author={Wang, Liang and Wang, Kezhi and Pan, Cunhua and Xu, Wei and Aslam, Nauman and Hanzo, Lajos},
  journal={IEEE Transactions on Cognitive Communications and Networking},
  volume={7},
  number={1},
  pages={73--84},
  year={2020},
  publisher={IEEE}
}

@book{bertsekas2019reinforcement,
  title={Reinforcement learning and optimal control},
  author={Bertsekas, Dimitri},
  volume={1},
  year={2019},
  publisher={Athena Scientific}
}

@article{bellman1957markovian,
  title={A Markovian decision process},
  author={Bellman, Richard},
  journal={Journal of mathematics and mechanics},
  pages={679--684},
  year={1957},
  publisher={JSTOR}
}

@article{kaelbling1998planning,
  title={Planning and acting in partially observable stochastic domains},
  author={Kaelbling, Leslie Pack and Littman, Michael L and Cassandra, Anthony R},
  journal={Artificial intelligence},
  volume={101},
  number={1-2},
  pages={99--134},
  year={1998},
  publisher={Elsevier}
}

@article{fang2024multi,
  title={Multi-UAV collaborative path planning based on multi-agent soft actor critic},
  author={Fang, CL and Yang, FS and Pan, Q},
  journal={Sci. Sin. Inf},
  volume={54},
  pages={1871--1883},
  year={2024}
}

@article{rashid2020monotonic,
  title={Monotonic value function factorisation for deep multi-agent reinforcement learning},
  author={Rashid, Tabish and Samvelyan, Mikayel and De Witt, Christian Schroeder and Farquhar, Gregory and Foerster, Jakob and Whiteson, Shimon},
  journal={Journal of Machine Learning Research},
  volume={21},
  number={178},
  pages={1--51},
  year={2020}
}

@article{han2025deep,
  title={A Deep Reinforcement Learning Method for Collision Avoidance with Dense Speed-Constrained Multi-UAV},
  author={Han, Jiale and Zhu, Yi and Yang, Jian},
  journal={IEEE Robotics and Automation Letters},
  year={2025},
  publisher={IEEE}
}

@inproceedings{sun2024t2mac,
  title={T2mac: Targeted and trusted multi-agent communication through selective engagement and evidence-driven integration},
  author={Sun, Chuxiong and Zang, Zehua and Li, Jiabao and Li, Jiangmeng and Xu, Xiao and Wang, Rui and Zheng, Changwen},
  booktitle={Proceedings of the AAAI Conference on Artificial Intelligence},
  volume={38},
  number={13},
  pages={15154--15163},
  year={2024}
}

@article{tampuu2017multiagent,
  title={Multiagent cooperation and competition with deep reinforcement learning},
  author={Tampuu, Ardi and Matiisen, Tambet and Kodelja, Dorian and Kuzovkin, Ilya and Korjus, Kristjan and Aru, Juhan and Aru, Jaan and Vicente, Raul},
  journal={PloS one},
  volume={12},
  number={4},
  pages={e0172395},
  year={2017},
  publisher={Public Library of Science San Francisco, CA USA}
}

@article{sunehag2017value,
  title={Value-decomposition networks for cooperative multi-agent learning},
  author={Sunehag, Peter and Lever, Guy and Gruslys, Audrunas and Czarnecki, Wojciech Marian and Zambaldi, Vinicius and Jaderberg, Max and Lanctot, Marc and Sonnerat, Nicolas and Leibo, Joel Z and Tuyls, Karl and others},
  journal={arXiv preprint arXiv:1706.05296},
  year={2017}
}

@inproceedings{son2019qtran,
  title={Qtran: Learning to factorize with transformation for cooperative multi-agent reinforcement learning},
  author={Son, Kyunghwan and Kim, Daewoo and Kang, Wan Ju and Hostallero, David Earl and Yi, Yung},
  booktitle={International conference on machine learning},
  pages={5887--5896},
  year={2019},
  organization={PMLR}
}

@article{ackermann2019reducing,
  title={Reducing overestimation bias in multi-agent domains using double centralized critics},
  author={Ackermann, Johannes and Gabler, Volker and Osa, Takayuki and Sugiyama, Masashi},
  journal={arXiv preprint arXiv:1910.01465},
  year={2019}
}

%%%%%%%%%%%%%%%%%%%%%%%%%%%%%%%%%%%%%%%%%%%%%%%%%%%%%%%%%%%%%%%%%%%%%%%%
%%% APPENDIX
%%%%%%%%%%%%%%%%%%%%%%%%%%%%%%%%%%%%%%%%%%%%%%%%%%%%%%%%%%%%%%%%%%%%%%%%
\onecolumn
\section*{APPENDIX}
\addcontentsline{toc}{section}{Appendix} 

\setcounter{section}{0}
\renewcommand{\thesection}{\Alph{section}}

\section{HYPERPARAMETER DETAILS}
\label{appendix:hyperparameters}

To ensure the reproducibility of our work, this section provides a detailed account of the hyperparameters used for the neural network architectures and the training process.

\subsection{Neural Network Architectures}
\label{appendix:nn_arch}

The architectural details of the actor and critic networks within the AC-MASAC framework are presented in Table \ref{tab:nn_architecture_concise_v1}. We utilize Multi-Layer Perceptrons (MLPs) with ReLU activation functions for all hidden layers. The final layer of the actor network employs a Tanh activation to constrain the output actions within the range [-1, 1].

\begin{table}[H]
\centering
\caption{Concise neural network architectures for AC-MASAC.}
\label{tab:nn_architecture_concise_v1}
\begin{tabular}{@{}ll@{}}
\toprule
\textbf{Network Component} & \textbf{Architecture Details} \\ \midrule
\textbf{Actor Network} & \\
\quad Encoder & Input(7) $\rightarrow$ Linear(256) $\rightarrow$ ReLU $\rightarrow$ Linear(128) \\
\quad Attention & Multi-Head (4 Heads, Dropout=0.1) \\
\quad Head & Context $\rightarrow$ Linear(128) $\rightarrow$ ReLU $\rightarrow$ Linear(2) $\rightarrow$ Tanh \\ \midrule
\textbf{Critic Network} & \\
\quad Encoder & Input(9) $\rightarrow$ Linear(256) $\rightarrow$ ReLU $\rightarrow$ Linear(128) \\
\quad Attention & Multi-Head (4 Heads, Dropout=0.1) \\
\quad Head (Q1 \& Q2) & Concat(Context, Action) $\rightarrow$ Linear(128) $\rightarrow$ ReLU $\rightarrow$ Linear(1) \\ \bottomrule
\end{tabular}
\end{table}

\subsection{Training Hyperparameters}
\label{appendix:training_hps}

The key hyperparameters used during the training of the AC-MASAC agent and all baseline models are summarized in Table \ref{tab:training_hps}. These parameters were selected based on common practices in deep reinforcement learning literature and were kept consistent across all experiments to ensure a fair comparison.

\begin{table}[H]
\centering
\caption{Training hyperparameters for all experiments.}
\label{tab:training_hps}
\begin{tabular}{@{}ll@{}}
\toprule
\textbf{Hyperparameter} & \textbf{Value} \\ \midrule
Optimizer & Adam \\
Learning Rate (Actor) & $1 \times 10^{-4}$ \\
Learning Rate (Critic) & $3 \times 10^{-4}$ \\
Learning Rate (Entropy) & $3 \times 10^{-4}$ \\
Discount Factor ($\gamma$) & 0.99 \\
Target Network Update Rate ($\tau$) & 0.01 \\
Replay Buffer Capacity & $50000$ \\
Batch Size & 256 \\
Target Entropy ($\mathcal{H}$) & -0.1 \\ \bottomrule
\end{tabular}
\end{table}

\section{CURRICULUM STAGE SPECIFICATIONS}
\label{appendix:curriculum}

\subsection{Environmental Configurations}
Our progressive curriculum learning framework utilizes a deterministic sequence of tasks with incrementally increasing complexity. The specific environmental configurations for each stage of the curriculum are detailed in Table \ref{tab:curriculum_stages}. The transition to the next stage is triggered upon meeting the holistic competency criteria outlined in the main text.
\begin{table}[H]
\centering
\caption{Environmental configurations for each curriculum learning stage.}
\label{tab:curriculum_stages}
\begin{tabular}{@{}cccccc@{}}
\toprule
\textbf{Stage} & \textbf{Leader Count} & \textbf{Follower Count} & \textbf{Obstacle Count} \\ \midrule
1 & 1 & 1 & 0  \\
2 & 1 & 2 & 1  \\
3 & 1 & 3 & 2  \\
4 & 1 & 4 & 2  \\
5 & 1 & 5 & 3  \\ \bottomrule
\end{tabular}
\end{table}

\subsection{Knowledge Transfer Mechanism}
Our parameter transfer strategy employs a differentiated approach to accommodate the distinct characteristics and functional requirements of various network components. For actor networks, the leader network parameters $\theta_L^{(t)}$ are completely transferred to the new stage as $\theta_L^{(t+1)} = \theta_L^{(t)}$, since the leader's decision-making patterns and coordination responsibilities remain consistent across curriculum stages. In contrast, follower networks adopt an expansive transfer strategy where existing follower network parameters $\{\theta_{F_i}^{(t)}\}_{i=1}^{N_t}$ are fully preserved to maintain learned collaborative behaviors, while newly added follower networks acquire foundational cooperation capabilities through parameter replication: $\psi_{F_j}^{(t+1)} = \psi_{F_{N_t}}^{(t)}$ for $j > N_t$, where $N_t$ and $N_{t+1}$ represent the number of followers in stages $t$ and $t+1$ respectively. The critic network employs a reinitialization strategy $\phi^{(t+1)} \sim \mathcal{N}(0, \sigma^2)$ due to changes in state-action space dimensionality from $\mathcal{S}^{(t)} \times \mathcal{A}^{(t)}$ to $\mathcal{S}^{(t+1)} \times \mathcal{A}^{(t+1)}$, preserving the network architecture while resetting all parameter weights and relying on new-stage experience data to rapidly relearn value function $Q_{\phi}(s,a)$. Entropy adjustment parameters $\{\alpha_L^{(t)}, \alpha_F^{(t)}\}$ are preserved as $\{\alpha_L^{(t+1)}, \alpha_F^{(t+1)}\} = \{\alpha_L^{(t)}, \alpha_F^{(t)}\}$, maintaining the optimal exploration-exploitation balance established in the previous stage. This differentiated parameter transfer strategy ensures effective knowledge inheritance while providing appropriate flexibility for new-stage learning, achieving an optimal balance between knowledge accumulation and adaptability throughout the curriculum learning process.
\subsection{Stage-Proportional Experience Sampling}
To mitigate catastrophic forgetting, the experience replay buffer $\mathcal{D}$ is managed as a collection of stage-specific sub-buffers. During training at stage $k$, a mini-batch is composed of samples from the current stage buffer $\mathcal{D}_k$ and a historical buffer $\mathcal{D}_{\text{hist}} = \bigcup_{j=1}^{k-1} \mathcal{D}_j$. The proportion of experiences sampled from the historical buffer, $\rho_{\text{old}}(k)$, is determined by the stage-dependent decay function detailed in Equation \ref{eq:ratio_symbolic}, using the hyperparameters specified in Section 3.3. This mechanism ensures a dynamic balance between adapting to the current task and retaining previously acquired knowledge.

\section{SUPPLEMENTARY RESULTS}
\label{appendix:extra_results}
This section provides additional qualitative and quantitative results to complement the analysis in the main paper, offering deeper insights into the AC-MASAC algorithm's performance characteristics and behavioral patterns.
\subsection{Trajectory Analysis in Challenging Scenarios}

\begin{figure}[H]
    \centering
    \begin{subfigure}{0.3\textwidth}
        \includegraphics[width=\linewidth]{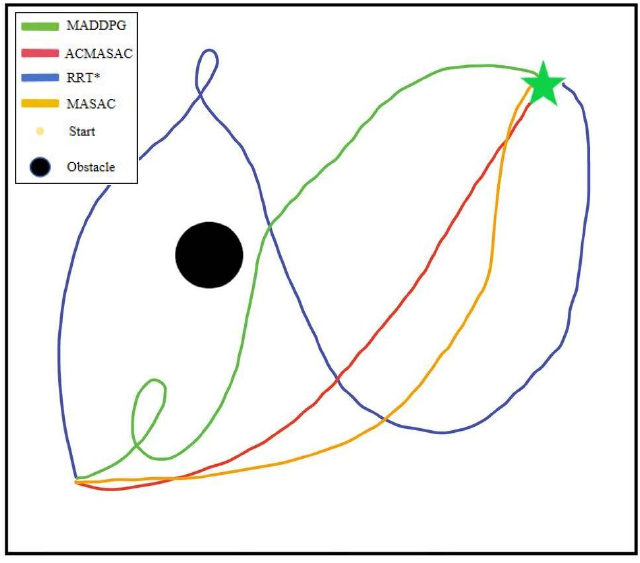} 
        \caption{Simple environment}
        \label{fig:traj_simple}
    \end{subfigure}
    \hfill
    \begin{subfigure}{0.3\textwidth}
        \includegraphics[width=\linewidth]{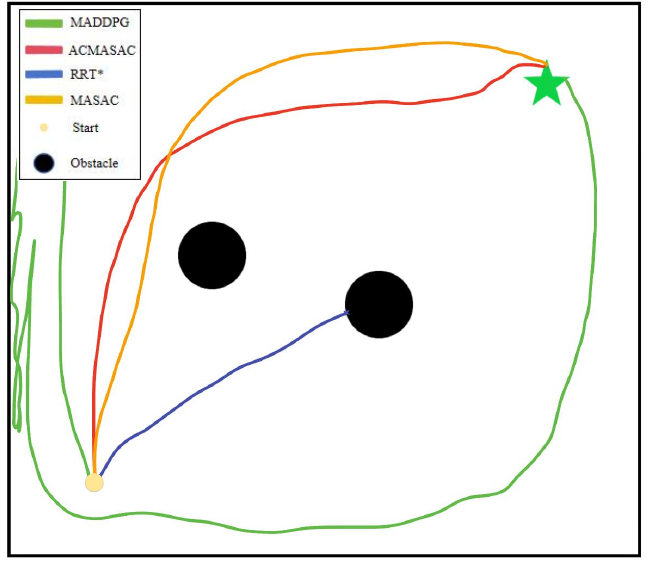} 
        \caption{Moderate environment}
        \label{fig:traj_moderate}
    \end{subfigure}
    \hfill
    \begin{subfigure}{0.3\textwidth}
        \includegraphics[width=\linewidth]{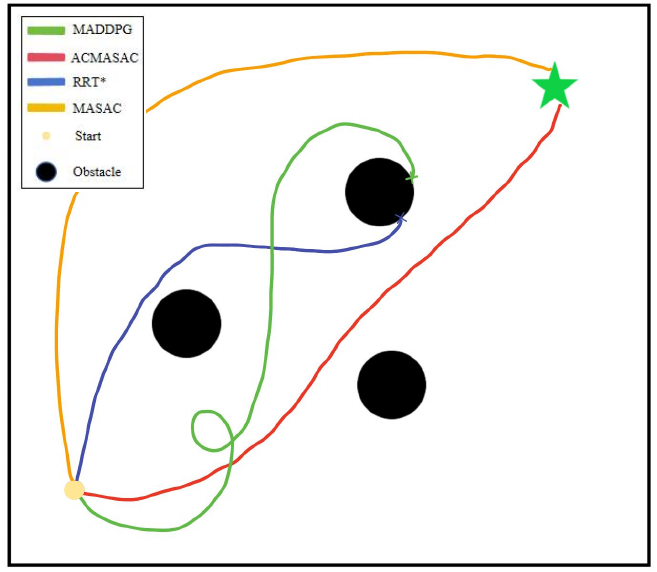} 
        \caption{Complex environment}
        \label{fig:traj_complex}
    \end{subfigure}
    \caption{Trajectory comparison across environments of varying complexity. AC-MASAC  demonstrates superior path efficiency and smoother navigation compared to baseline methods including MADDPG, RRT*, and MASAC. Black circles represent obstacles, green stars indicate target positions.}
    \label{fig:appendix_trajectories}
\end{figure}

Figure \ref{fig:appendix_trajectories} illustrates typical agent trajectories for AC-MASAC and baseline algorithms across various challenging test scenarios. These visualizations demonstrate the superior path planning capabilities and obstacle avoidance strategies of the proposed method.

The trajectory analysis reveals several key advantages of AC-MASAC: (1) \textbf{Path Efficiency}: AC-MASAC consistently generates shorter and more direct paths to targets across all complexity levels; (2) \textbf{Smooth Navigation}: The trajectories exhibit reduced oscillations and smoother curves, indicating better control stability; (3) \textbf{Obstacle Avoidance}: More effective navigation around obstacles with improved safety margins compared to baseline methods.
\subsection{Formation Dynamics Analysis}

\begin{figure}[H]
    \centering
    \begin{subfigure}{0.32\textwidth}
        \includegraphics[width=\linewidth]{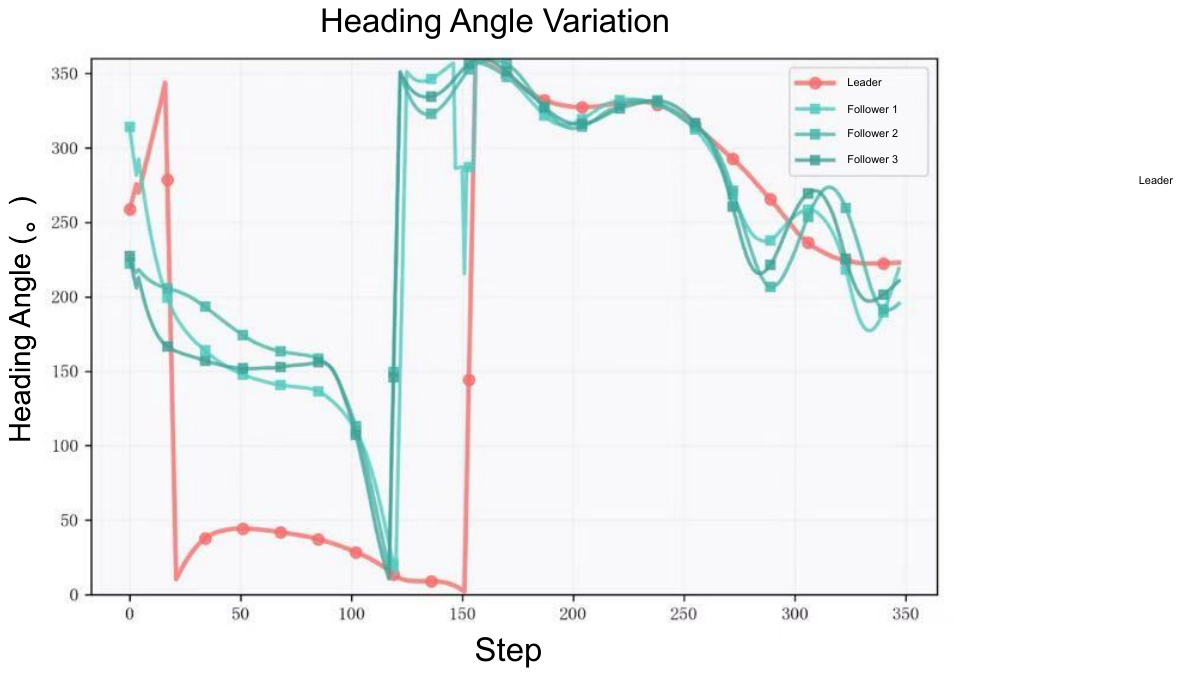} 
        \caption{Heading angle variation}
        \label{fig:heading_analysis}
    \end{subfigure}
    \hfill
    \begin{subfigure}{0.32\textwidth}
        \includegraphics[width=\linewidth]{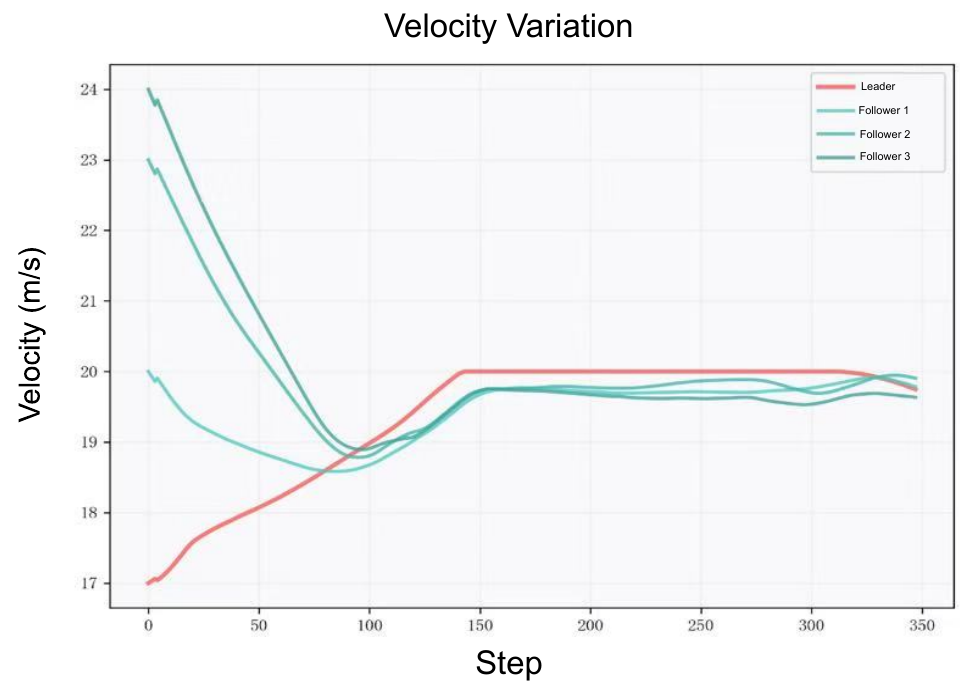}
        \caption{Velocity coordination}
        \label{fig:velocity_analysis}
    \end{subfigure}
    \hfill
    \begin{subfigure}{0.30\textwidth}
        \includegraphics[width=\linewidth]{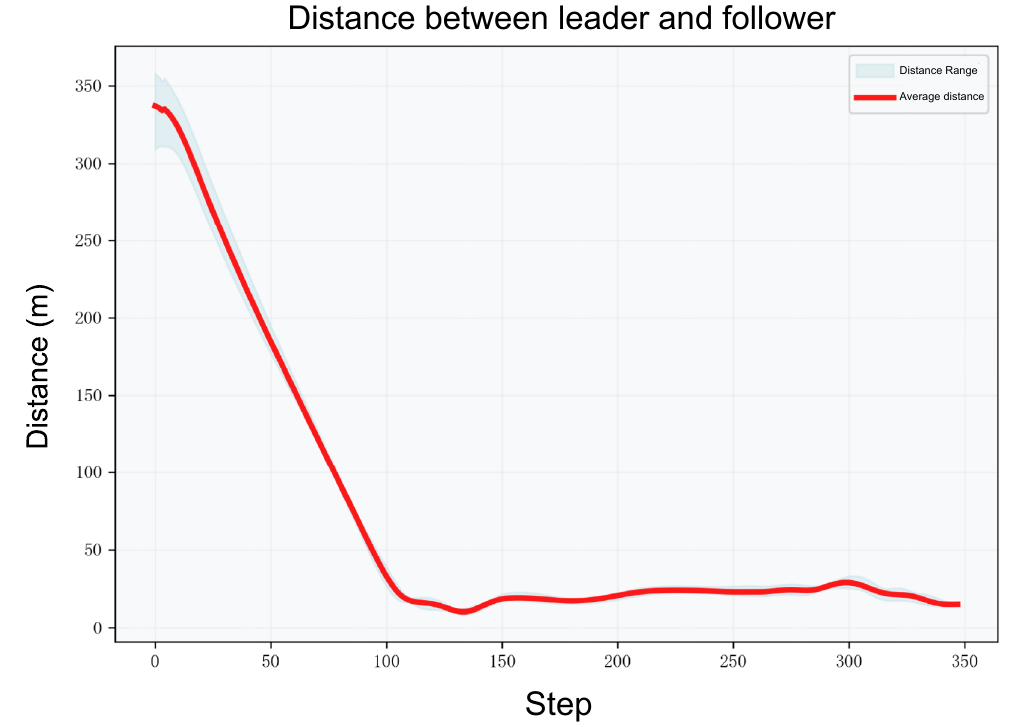} 
        \caption{Distance evolution}
        \label{fig:distance_analysis}                           
    \end{subfigure}
    \caption{Formation dynamics analysis showing multi-agent coordination performance over time. The plots demonstrate the effectiveness of the attention mechanism in achieving synchronized behavior among heterogeneous agents.}
    \label{fig:formation_dynamics}
\end{figure}

Figure \ref{fig:formation_dynamics} presents detailed analysis of the formation control capabilities, examining heading angle coordination, velocity synchronization, and inter-agent distance management.

The formation dynamics analysis reveals the effectiveness of AC-MASAC's attention-based coordination mechanism across three critical aspects. In heading angle coordination (Figure~\ref{fig:heading_analysis}), the leader and followers initially exhibit significant heading disparities, but through the attention mechanism, followers progressively align with the leader's heading direction, achieving synchronization within approximately 100 time steps and maintaining alignment with deviations below $5°$ in the final phase. Simultaneously, velocity synchronization (Figure~\ref{fig:velocity_analysis}) demonstrates rapid adaptation capabilities, where the leader's velocity changes from $17~\text{m/s}$ to $19~\text{m/s}$ are accurately tracked by followers, with $95\%$ synchronization achieved within 150 time steps, validating the effectiveness of role-specific actor networks in maintaining coordinated movement. Furthermore, distance management (Figure~\ref{fig:distance_analysis}) shows rapid convergence from initial separation ($\approx 350~\text{m}$) to stable formation distance ($20$--$30~\text{m}$) within 100 time steps, with subsequent maintenance of consistent spacing demonstrating robust formation-keeping capabilities. These coordinated behaviors collectively validate the superior multi-agent coordination performance achieved through the proposed attention mechanism and heterogeneous network architecture.
%%%%%%%%%%%%%%%%%%%%%%%%%%%%%%%%%%%%%%%%%%%%%%%%%%%%%%%%%%%%%%%%%%%%%%%%

\end{document}